\documentclass[11pt,letterpaper]{article}

\usepackage[letterpaper,margin=1in]{geometry}
\usepackage[T1]{fontenc}
\usepackage[utf8]{inputenc}
\usepackage{lmodern}
\usepackage{microtype}
\usepackage{amsmath,amssymb}

\usepackage{graphicx}
\graphicspath{{figures/}}
\usepackage{booktabs}
\usepackage{array}
\usepackage{longtable}
\usepackage{xcolor}
\definecolor{NavyBlue}{HTML}{1F4E79}

\usepackage{float}
\usepackage[font=small,labelfont=bf,labelsep=period]{caption}

\usepackage[colorlinks=true,linkcolor=NavyBlue,citecolor=NavyBlue,urlcolor=NavyBlue,
            breaklinks=true]{hyperref}
\usepackage{url}
\usepackage[round,numbers,sort&compress]{natbib}
\DeclareUnicodeCharacter{2212}{-}
\DeclareUnicodeCharacter{2192}{$\rightarrow$}
\DeclareUnicodeCharacter{2264}{$\leq$}
\DeclareUnicodeCharacter{2265}{$\geq$}
\DeclareUnicodeCharacter{2248}{$\approx$}
\DeclareUnicodeCharacter{00D7}{$\times$}
\DeclareUnicodeCharacter{00B7}{$\cdot$}
\DeclareUnicodeCharacter{2013}{--}
\DeclareUnicodeCharacter{2018}{`}
\DeclareUnicodeCharacter{2019}{'}
\DeclareUnicodeCharacter{201C}{``}
\DeclareUnicodeCharacter{201D}{''}
\DeclareUnicodeCharacter{2026}{\ldots}
\DeclareUnicodeCharacter{2022}{$\bullet$}

\providecommand{\tightlist}{\setlength{\itemsep}{0pt}\setlength{\parskip}{0pt}}

\title{\textbf{OncoTraj: a public benchmark for longitudinal resistance
prediction in EGFR-mutant non-small-cell lung cancer on osimertinib}}

\author{
  Abhijoy Sarkar\thanks{Corresponding author. \href{mailto:abhijoy.sar@gmail.com}{abhijoy.sar@gmail.com}}\\
  Span AI
  \and
  Aarchi Singh Thakur\thanks{Corresponding author. \href{mailto:aarchisingh.t@gmail.com}{aarchisingh.t@gmail.com}}\\
  Span AI
}

\date{May 2026}

\begin{document}
\maketitle

\begin{abstract}
\noindent
Resistance to first-line osimertinib in EGFR-mutant non-small-cell lung cancer
(NSCLC) is the canonical example of predictable clonal evolution under
therapeutic pressure, yet no public benchmark exists for training or evaluating
computational models on the corresponding longitudinal patient trajectories.
We introduce \textbf{OncoTraj}, a public benchmark of \textbf{813 EGFR-mutant
NSCLC patients} receiving first-line osimertinib, harmonized from three
real-world clinical-genomic sources: MSK-CHORD (672 patients), the AACR Project
GENIE Biopharma Collaborative NSCLC cohort (34 patients), and the published
FLAURA molecular-resistance supplement (107 patients). OncoTraj defines three
locked tasks: (A) binary classification of progression by a fixed 12-month
landmark, (B) regression of time-to-first-progression in days, and (C)
six-class classification of the dominant resistance mechanism at progression.
We release the harmonized dataset under a permissive schema, patient-level
train/validation/test splits with an audited no-leakage guarantee, an
open-source evaluation harness, and six reference baselines spanning a
majority-class predictor, logistic regression, random forest, XGBoost, an
LSTM, and a multi-task transformer.

OncoTraj's contribution is to establish a public floor and a measured ceiling for this problem, and to identify the binding constraint behind that ceiling. With v1's single-timepoint snapshot features, \textbf{no task clears chance on clean within-source evaluation}, and the uniformity of that ceiling across every model class localizes the limit to the \textbf{input modality} --- single-snapshot tissue NGS rather than serial ctDNA trajectories --- not the algorithm. The evidence: of the three tasks, time-to-progression (Task B) is the only one with any above-floor signal, yet on a within-source (MSK-CHORD-only) hold-out the best mean-absolute-error is 293 days against a 303-day majority floor (a $\sim$10-day gap with almost fully overlapping confidence intervals) and the survival ranking is at chance (C-index 0.43--0.49). The higher mixed-source figures --- landmark-progression ROC-AUC 0.680 and Task B 252-day MAE / 0.66 concordance --- are inflated by a dataset-membership confound: the date-less FLAURA subset has a structurally fixed label that a source-aware model predicts exactly. A v1.1 distribution-shift analysis (train on MSK-CHORD, test on held-out sources, with an added Cox proportional-hazards baseline) confirms the ceiling: cross-source point prediction collapses below the majority floor, and only a weak Cox hazard-ranking signal persists (C-index 0.597, 95\% CI 0.45--0.75 --- suggestive, not significant). Per-mechanism classification fails outright because tissue next-generation sequencing under-captures the dominant on-target resistance mutation (EGFR C797S), which requires serial ctDNA to detect. The benchmark does, however, recover a reproducible, literature-consistent prognostic association: TP53 co-mutation raises the 12-month progression rate from 29\% to 59\% cohort-wide, with the same direction preserved within MSK-CHORD and GENIE BPC independently --- confirming known osimertinib biology, though it does not by itself yield above-chance within-source discrimination. Taken together, OncoTraj establishes a real, reproducible, leakage-audited baseline and converts the modality limit into concrete design requirements for a serial-ctDNA-enriched v2.

\vspace{0.4em}

\vspace{0.4em}
\noindent\textbf{License.} MIT (code), CC-BY 4.0 (dataset card).\quad
\textbf{Target server.} bioRxiv.
\end{abstract}

\vspace{0.5em}
\hypertarget{introduction}{%
\section{Introduction}\label{introduction}}

EGFR-mutant non-small-cell lung cancer (NSCLC) accounts for roughly 15\%
of newly-diagnosed lung adenocarcinomas in Western populations and a
substantially higher fraction in East and South Asian populations
\cite{siegel2023cancerstats}. Patients with sensitizing EGFR mutations,
most commonly exon 19 deletions (\texttt{exon19del}) and the L858R
missense, receive osimertinib, a third-generation tyrosine-kinase
inhibitor (TKI), as standard-of-care first-line therapy
\cite{soria2018flaura,ramalingam2020flauraos}. Median progression-free
survival on first-line osimertinib in the FLAURA trial was 18.9 months;
almost every patient progresses within two to three years.

The pathways to progression are characterized to an unusual degree. In
the first-line setting, acquired resistance is molecularly
heterogeneous: the most frequently identified mechanisms are MET
amplification (\textasciitilde16\%) and the on-target EGFR C797S
mutation (\textasciitilde6\%), with EGFR or HER2 amplification and
histologic transformation to small-cell or squamous lineage accounting
for smaller fractions and a substantial share of patients having no
single identifiable mechanism. Notably, the EGFR T790M mutation that
dominates \emph{second-line} resistance is essentially absent after
first-line osimertinib \cite{chmielecki2023,mok2017aura3}. Importantly,
serial circulating-tumor DNA (ctDNA) monitoring during osimertinib
therapy is prognostic, early plasma EGFR-mutation clearance (by week
3--6) predicts longer progression-free survival, and dynamic ctDNA
changes track the emergence of resistant clones \cite{gray2023ctdna}.
The signal exists. The clinical question is whether it can be read
systematically.

This setting, a single tumour type, a single first-line agent, a
well-characterised set of resistance trajectories, and an existing
serial-ctDNA monitoring infrastructure, is the right testbed for
computational models of longitudinal patient state. The intellectual
template is the TRACERx programme (Swanton and colleagues), which
established that NSCLC evolution under selective pressure is partially
predictable and that serial ctDNA can track clonal dynamics through
treatment \cite{jamalhanjani2017tracerx,frankell2023tracerxevo}.
TRACERx, however, was built on early-stage \emph{resectable} disease
followed to surgical relapse; the analogous resource for \emph{advanced}
EGFR-mutant disease on first-line osimertinib , the setting where
resistance prediction is clinically actionable, does not exist as a
public benchmark. Individual trial supplements include longitudinal data
on dozens to a few hundred patients each, but the data formats are
heterogeneous, splits are not standardised, and there is no shared
evaluation harness. Researchers who wish to model resistance prediction
on osimertinib must either reconstruct cohorts from scratch or work on
private hospital data that cannot be released.

We therefore introduce \textbf{OncoTraj}, a public benchmark assembled
from three real-world clinical-genomic sources: (i) the MSK-CHORD
real-world dataset \cite{jee2024mskchord} (672 patients), (ii) the AACR
Project GENIE Biopharma Collaborative NSCLC cohort
\cite{choudhury2023geniebpc} (34 patients), and (iii) the FLAURA
molecular-resistance supplement of Chmielecki et al.~2023
\cite{chmielecki2023} (107 patients). The harmonized v1 cohort comprises
\textbf{813 patients} with EGFR-mutant NSCLC receiving first-line
osimertinib. Patients from MSK-CHORD and GENIE BPC whose first EGFR-TKI
was not osimertinib are ingested through the same pipeline but held out
as a prior-TKI sensitivity slice, keeping the benchmark cohort strictly
first-line. We define three locked prediction tasks, release
patient-level train/validation/test splits with an audited no-leakage
guarantee, ship an open-source evaluation package, and report results
for six reference baselines spanning classical machine learning and
small deep models. Beyond the benchmark itself, the harmonized cohort
recovers a reproducible prognostic association consistent with the
published osimertinib-resistance literature --- TP53 co-mutation raises the 12-month progression rate from 29\% to 59\% cohort-wide, with the same direction preserved within MSK-CHORD and GENIE BPC independently --- evidence that the benchmark surfaces real biology even where snapshot features do not support above-chance
prediction.

The paper's contributions are:

\begin{enumerate}
\def\labelenumi{\arabic{enumi}.}
\tightlist
\item
  \textbf{A harmonized 813-patient public dataset and schema}
  (\texttt{oncotraj\ v1}) for longitudinal resistance prediction in
  EGFR-mutant NSCLC on osimertinib, spanning three real clinical-genomic
  sources.
\item
  \textbf{A locked, leakage-audited evaluation protocol} with
  patient-level splits, defined metrics, a shipped feature-leakage test,
  and a published leaderboard format.
\item
  \textbf{Six reference baselines} with full source code, trained
  weights, and reproducible evaluation reports, evaluated with bootstrap
  confidence intervals.
\item
  \textbf{A candid error analysis} that documents where the baselines
  fail, why, and what an incoming v2 dataset must contain to address
  each failure mode, including a full source-flag confound analysis for
  the landmark task and a conversion of the one structural negative
  result (mechanism classification) into a concrete dataset-design
  requirement.
\end{enumerate}

\textbf{What this benchmark enables.} OncoTraj v1 fixes the splits, schema, and leakage audit that longitudinal-resistance methods will need, so that methods designed for serial molecular dynamics --- for example change-point detectors operating on ctDNA detection-indicator sequences --- can be evaluated head-to-head on OncoTraj v2 once serial samples are available. v1 deliberately measures the ceiling of the single-snapshot regime; v2 is where methods built for the serial regime can demonstrate they beat it.

We are deliberate about what this paper does \textbf{not} claim. We do
not claim clinical utility. We do not claim deep learning works better
than classical baselines at the v1 cohort size, our results show the
opposite. We do not claim mechanism-class prediction (Task C) is solved
at any defensible level. And we do not present the headline
12-month-landmark discrimination as a within-source effect: §5 reports
both the full-cohort number and the more conservative within-source
estimate, and documents the cross-source structure that separates them.
Section 5 documents all of these limitations in detail.

\begin{figure}[t]
\hypertarget{fig:schematic}{%
\centering
\includegraphics[width=1\linewidth]{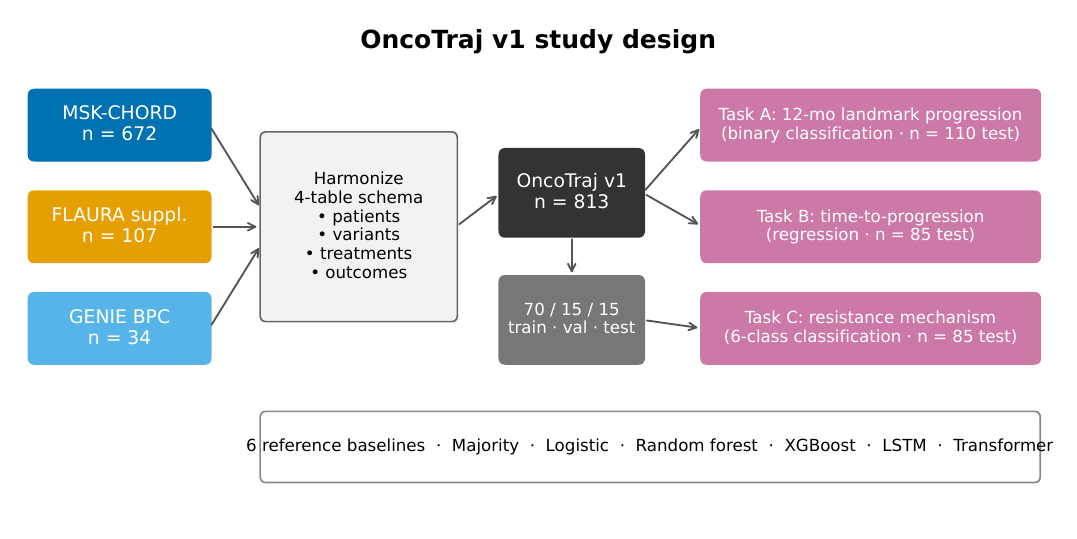}
\caption{OncoTraj v1 study design: three real-world clinical-genomic
sources are harmonized into a unified four-table schema, producing the
813-patient first-line-osimertinib cohort. Patient-level 70/15/15 splits
feed the three locked tasks (A, B, C) against six reference
baselines.}\label{fig:schematic}
}
\end{figure}

\hypertarget{related-work}{%
\section{Related work}\label{related-work}}

\textbf{Clonal-evolution-aware resistance modelling.} The TRACERx
programme (Swanton and colleagues) established that NSCLC tumours under
therapeutic pressure follow partially convergent evolutionary paths,
with recurring driver--passenger hierarchies, and that serial ctDNA can
track clonal dynamics and metastatic dissemination through treatment
\cite{jamalhanjani2017tracerx,frankell2023tracerxevo}. That body of work
was conducted on early-stage resectable disease followed to surgical
relapse, a different clinical setting and endpoint (relapse-free
survival) than ours. OncoTraj deliberately transplants the TRACERx
clonal-tracking logic into the \emph{advanced, first-line-osimertinib}
setting, where the endpoint is time-to-progression on a targeted agent
and the dominant resistance mechanisms (EGFR C797S, MET amplification)
are pharmacologically actionable. Exploratory analyses of the FLAURA and
AURA3 trials show that serial ctDNA, specifically early plasma
EGFR-mutation clearance, predicts progression-free survival on
osimertinib \cite{gray2023ctdna}. More recent serial-ctDNA analyses on
osimertinib-based regimens reinforce the same pattern, with early plasma
EGFR-mutation clearance predicting longer PFS and acquired EGFR C797S
detected dynamically in 13--22\% of patients on osimertinib in the
ETOP-BOOSTER trial \cite{soo2024etopbooster} and tracked in single-arm
cohorts under intensified osimertinib regimens
\cite{zwierenga2025ctdna}. Several groups have published single-cohort
models of resistance timing or mechanism on EGFR-mutant NSCLC, primarily
within trial populations, without a shared evaluation framework or
public dataset release. Recent biologically-informed neural networks for
drug-response prediction on EGFR-pathway cancer cell-line panels also
surface TP53 and KRAS as top biomarkers of osimertinib resistance,
consistent with the co-mutation signal we report in \S5.2
\cite{ran2025knet}, the gap OncoTraj fills.

\textbf{Benchmarks in clinical machine learning.} ImageNet
\cite{deng2009imagenet} and CASP \cite{moult1995casp} anchored
breakthroughs in computer vision and protein-structure prediction
respectively, both by providing a public benchmark against which methods
could be compared. Clinical machine learning has fewer such anchors.
MIMIC-III/IV \cite{johnson2023mimiciv} is the closest large-scale
analogue for general ICU and EHR data; the oncology equivalent does not
exist at the level of granularity needed for resistance prediction.
OncoTraj aims to fill that gap for one specific failure mode, with the
explicit understanding that v1's n=813 is a floor and that future
versions will expand both depth (more samples per patient, especially
serial ctDNA) and breadth (more cancer types).

\textbf{Multi-modal trajectory models in medicine.} Transformer-style
architectures over irregularly-sampled clinical events have shown
promise in EHR-based mortality and readmission prediction
\cite{choi2016doctorai}. We adapt this template to the
resistance-prediction setting, but our results show that at v1's cohort
size deep models are not yet a productive over-investment relative to
classical baselines.

\hypertarget{dataset}{%
\section{Dataset}\label{dataset}}

\hypertarget{cohort-construction}{%
\subsection{Cohort construction}\label{cohort-construction}}

We harmonize patients from three real-world clinical-genomic sources
into a unified four-table schema (\texttt{patients}, \texttt{variants},
\texttt{treatments}, \texttt{outcomes}) versioned as
\texttt{oncotraj-schema/1.0.0}. The detailed schema, including field
types, allowed enumerations, and missingness conventions, is in
\texttt{DATASET\_SPEC.md} in the public repository.

Inclusion criteria for v1:

\begin{enumerate}
\def\labelenumi{\arabic{enumi}.}
\tightlist
\item
  Histologically confirmed NSCLC (lung adenocarcinoma or NSCLC-NOS).
\item
  Documented EGFR sensitizing variant (exon 19 deletion, L858R, G719X,
  L861Q, S768I, or compound sensitizing).
\item
  Osimertinib as the patient's \textbf{first} EGFR tyrosine-kinase
  inhibitor (1L-osimertinib). Patients whose first EGFR-TKI was
  erlotinib, gefitinib, afatinib, or dacomitinib are excluded from the
  v1 cohort and retained as a held-out prior-TKI sensitivity slice.
\item
  At least one molecular sample (tissue NGS or ctDNA) before or at
  osimertinib initiation.
\end{enumerate}

Patients meeting these criteria are retained regardless of follow-up
duration or progression status, with censoring handled at the task
level.

The harmonized v1 cohort comprises \textbf{813 patients} across three
real sources:

\begin{longtable}[]{@{}llll@{}}
\toprule
\begin{minipage}[b]{0.22\columnwidth}\raggedright
Source\strut
\end{minipage} & \begin{minipage}[b]{0.22\columnwidth}\raggedright
n (1L-osi cohort)\strut
\end{minipage} & \begin{minipage}[b]{0.22\columnwidth}\raggedright
held-out (prior-TKI / non-sensitizing)\strut
\end{minipage} & \begin{minipage}[b]{0.22\columnwidth}\raggedright
Progression data\strut
\end{minipage}\tabularnewline
\midrule
\endhead
\begin{minipage}[t]{0.22\columnwidth}\raggedright
MSK-CHORD (Jee et al.~2024)\strut
\end{minipage} & \begin{minipage}[t]{0.22\columnwidth}\raggedright
672\strut
\end{minipage} & \begin{minipage}[t]{0.22\columnwidth}\raggedright
289 / 98\strut
\end{minipage} & \begin{minipage}[t]{0.22\columnwidth}\raggedright
NLP-derived progression timeline (radiology impressions)\strut
\end{minipage}\tabularnewline
\begin{minipage}[t]{0.22\columnwidth}\raggedright
FLAURA supplement (Chmielecki et al.~2023)\strut
\end{minipage} & \begin{minipage}[t]{0.22\columnwidth}\raggedright
107\strut
\end{minipage} & \begin{minipage}[t]{0.22\columnwidth}\raggedright
,\strut
\end{minipage} & \begin{minipage}[t]{0.22\columnwidth}\raggedright
baseline + discontinuation samples\strut
\end{minipage}\tabularnewline
\begin{minipage}[t]{0.22\columnwidth}\raggedright
AACR GENIE BPC NSCLC v2.0 (Choudhury et al.~2023)\strut
\end{minipage} & \begin{minipage}[t]{0.22\columnwidth}\raggedright
34\strut
\end{minipage} & \begin{minipage}[t]{0.22\columnwidth}\raggedright
114 / 10\strut
\end{minipage} & \begin{minipage}[t]{0.22\columnwidth}\raggedright
abstractor-curated PFS endpoints (imaging + med-onc notes)\strut
\end{minipage}\tabularnewline
\bottomrule
\end{longtable}

GENIE BPC's 1L-osimertinib yield is modest because its enrollment window
predates the FLAURA-era shift of osimertinib to first line; most
osimertinib use in BPC is second-line or later, captured in the held-out
prior-TKI slice. We nonetheless ingest BPC for source diversity and
because its curated progression endpoints are higher quality than
NLP-derived events.

The EGFR-variant-class distribution across the v1 cohort: exon19del 424 (52\%), L858R 290 (36\%), compound\_with\_resistance 34, compound\_sensitizing 22, G719X 20, L861Q 18, exon20ins 4, S768I 1. 565
of 813 patients (69\%) have at least one documented progression event
with real timing.

A note on data provenance: every patient in the v1 cohort is real. An
earlier internal release used synthetic fixtures to validate the
ingestion pipeline end-to-end before controlled-access credentials were
obtained; those fixtures were replaced in their entirety once MSK-CHORD
and GENIE BPC access was granted, and no synthetic patient enters any
reported number.

\begin{figure}[t]
\hypertarget{fig:cohort}{%
\centering
\includegraphics[width=1\linewidth]{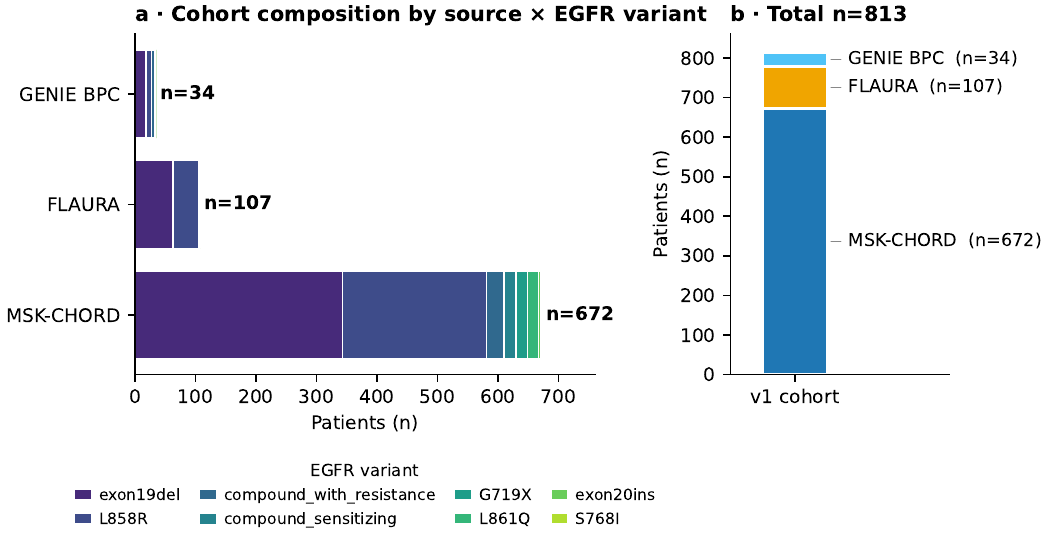}
\caption{OncoTraj v1 cohort composition. (a) Patient counts by source
and EGFR variant class. (b) Source-level totals across the 813-patient
v1 cohort.}\label{fig:cohort}
}
\end{figure}

\hypertarget{tasks}{%
\subsection{Tasks}\label{tasks}}

OncoTraj v1 defines three locked prediction tasks. The detailed
labelling rules are in \nolinkurl{LABELING_GUIDELINES.md} and were
authored before any model was trained.

\begin{itemize}
\tightlist
\item
  \textbf{Task A, 12-month landmark progression.} For each patient,
  given features computed strictly from events on or before osimertinib
  initiation, predict whether RECIST progression occurs within a fixed
  365-day landmark. The label is positive if a progression event falls
  at or before day 365, negative if the patient is confirmed
  progression-free at the landmark (followed past day 365 without
  progression), and the patient is \textbf{excluded} if censored before
  day 365 (insufficient follow-up to determine the label). This converts
  the degenerate ``ever-progresses'' form, near-useless on a cohort in
  which essentially every patient eventually progresses, into a
  balanced, clinically meaningful fixed-horizon target (cohort positive
  rate 45\%). Of the 813 patients, 760 have a determinable landmark
  label; 53 are censored before the landmark and dropped. We report
  ROC-AUC, Brier score, and Expected Calibration Error (ECE) on the test
  split. 
\item
  \textbf{Task B, Time-to-progression regression.} For each patient with
  a documented progression event, predict the time in days from
  osimertinib initiation to first RECIST progression. We report mean
  absolute error (MAE), root mean squared error (RMSE), and concordance
  index (C-index).
\item
  \textbf{Task C, Dominant resistance mechanism classification.} For
  each patient with a progression event, predict the dominant mechanism:
  EGFR\_C797S, MET\_amplification, EGFR\_amplification,
  histologic\_transformation, other, or unknown. We report macro-F1,
  weighted-F1, and per-class F1.
\end{itemize}

The complete mechanism-assignment rules, including hierarchy for
co-occurring mechanisms, low-VAF thresholds, and audit-log requirements,
are formalized in \nolinkurl{LABELING_GUIDELINES.md}.

\hypertarget{splits}{%
\subsection{Splits}\label{splits}}

We construct stratified patient-level splits with a 70/15/15
train/validation/test ratio. Stratification is on (i) source dataset and
(ii) Task A label, jointly, to ensure each source contributes
proportionally to each split and the test set does not become trivially
predictable from a source flag alone. The split manifest is shipped as
\texttt{\_splits.json} and is frozen at v1 release. Patient-level
splitting ensures no patient's trajectory is divided across splits.

The resulting splits: train n=570, validation n=121, test n=122. The
12-month landmark task (Task A) drops patients censored before the
landmark, leaving an effective test set of n=110; Task B and Task C are
evaluated on the subsets with a progression event / mechanism label
(test n=85). We report all headline results with bootstrap 95\%
confidence intervals (1000 resamples of the test set) so that the
precision of each estimate is explicit.

\hypertarget{methods}{%
\section{Methods}\label{methods}}

\hypertarget{features}{%
\subsection{Features}\label{features}}

For classical baselines we compute trajectory-aware patient-level
features in three groups:

\begin{itemize}
\tightlist
\item
  \textbf{Molecular dynamics.} \texttt{latest\_vaf}, maximum variant
  allele fraction in the patient's most recent variant-record sample;
  \texttt{vaf\_slope}, per-day change in EGFR sensitizing-variant VAF
  between earliest and latest sample, zero for single-sample patients;
  \texttt{variant\_burden}, total variant count per patient.
\item
  \textbf{Co-mutation profile.} Binary flags for co-occurring
  alterations in nine genes with established or candidate prognostic
  value on osimertinib (\texttt{TP53}, \texttt{RB1}, \texttt{CDKN2A},
  \texttt{PIK3CA}, \texttt{CTNNB1}, \texttt{MET}, \texttt{ERBB2},
  \texttt{KRAS}, \texttt{BRAF}), plus \texttt{comut\_burden}, the count
  of distinct non-EGFR mutated genes. TP53 in particular is the
  established osimertinib poor-prognosis co-mutation and is the dominant
  driver of the Task A signal (§5).
\item
  \textbf{Clinical.} \texttt{time\_on\_therapy\_days}, calendar days
  from osimertinib start to treatment end or last follow-up, clamped to
  zero for missing values.
\end{itemize}

To these we append one-hot encodings of \texttt{source\_dataset} and
\texttt{egfr\_variant\_class}. Every feature, including the co-mutation
flags, is computed strictly from records dated on or before the
patient's prediction cut-off (osimertinib initiation), enforced by an
automated test (\nolinkurl{tests/test_taska_feature_leakage.py}). We
discovered an initial label-leakage issue in pre-release Task A
features, see §6 for details, and corrected it before any reported
result. Because the \texttt{source\_dataset} one-hots can leak cohort
membership into the landmark label (§5.1), we report the headline Task A
result with source flags \textbf{removed}.

\hypertarget{baseline-models}{%
\subsection{Baseline models}\label{baseline-models}}

We report results for six baselines, all available in
\texttt{src/oncotraj/models/}:

\begin{itemize}
\tightlist
\item
  \textbf{Majority.} Predicts the majority class for Tasks A and C and
  the training-set mean target for Task B.
\item
  \textbf{Logistic regression.} Standard scikit-learn
  \texttt{LogisticRegression(class\_weight=\textquotesingle{}balanced\textquotesingle{})}
  for Tasks A and C; not applicable to Task B regression.
\item
  \textbf{Random forest.}
  \texttt{RandomForestClassifier}\slash\texttt{RandomForestRegressor} with
  500 trees, \texttt{max\_depth=8}, balanced class weights where
  applicable.
\item
  \textbf{XGBoost.} \texttt{XGBClassifier}/\texttt{XGBRegressor} with
  conservative defaults (200 boosting rounds,
  \texttt{learning\_rate=0.05}, \texttt{max\_depth=6}). Class weights
  enabled for Tasks A and C.
\item
  \textbf{LSTM.} A two-layer LSTM over typed clinical events; hidden
  sizes searched over \{32, 64, 128\} with learning rates \{1e-3,
  1e-4\}. Multi-task heads for Tasks B and C; Task A is read from the
  binary side of Task B's progression label. Targets for Task B are
  \texttt{log1p}-transformed before MSE loss and inverse-transformed
  (\texttt{expm1}) before metric computation.
\item
  \textbf{Transformer.} A small (10M-parameter) multi-task transformer
  over typed clinical events with rotary position embeddings on
  timestamp deltas. Three configurations were trained
  (\texttt{d\_model\ =\ 128,\ 256,\ 384}). Same multi-task head
  structure and target transform as LSTM.
\end{itemize}

All deep models train on Apple M-series hardware via MPS in under one
hour per configuration. No external accelerators were used. All baseline
runs are tracked in MLflow and reproducible from the public repository.

\hypertarget{evaluation}{%
\subsection{Evaluation}\label{evaluation}}

The evaluation package \texttt{oncotraj.eval} provides:

\begin{itemize}
\tightlist
\item
  A \texttt{predictions.csv} schema that submitters write to
  (\texttt{patient\_id,\ task\_a\_pred,\ task\_b\_pred,\ \ \ task\_c\_pred,\ task\_a\_proba,\ task\_c\_proba,\ split}).
\item
  A CLI entry point
  \texttt{oncotraj-eval\ -\/-predictions\ \textless{}csv\textgreater{}\ -\/-split\ test}
  that produces a versioned JSON evaluation report.
\item
  An auto-generated leaderboard markdown file kept in sync with the
  contents of \texttt{eval\_reports/}.
\end{itemize}

All metrics in this paper are computed by this package. We report on the
held-out test split throughout unless otherwise specified.

\hypertarget{results}{%
\section{Results}\label{results}}

\hypertarget{headline-results}{%
\subsection{Headline results}\label{headline-results}}

Table 1 reports headline metrics for the baselines across the three
tasks on the held-out test split (Task A n=110, Tasks B/C n=85), with
bootstrap 95\% confidence intervals on the two primary metrics (Task A
AUC, Task B MAE).

\textbf{Table 1.} Test-split performance on OncoTraj v1 (813-patient
cohort). Task A (12-month landmark) is evaluated on the n=110 test
patients with a determinable landmark label; Task B and Task C on the
n=85 test patients with a progression event / mechanism label. Task A is
reported \textbf{with source flags removed} (confound-free; see §5.2).
Bootstrap 95\% CIs from 1000 test-set resamples. Best non-trivial value
per metric in \textbf{bold}.

\begin{longtable}[]{@{}lllllll@{}}
\toprule
\begin{minipage}[b]{0.12\columnwidth}\raggedright
Model\strut
\end{minipage} & \begin{minipage}[b]{0.12\columnwidth}\raggedright
Task A · AUC {[}95\% CI{]}\strut
\end{minipage} & \begin{minipage}[b]{0.12\columnwidth}\raggedright
Task A · Brier\strut
\end{minipage} & \begin{minipage}[b]{0.12\columnwidth}\raggedright
Task A · ECE\strut
\end{minipage} & \begin{minipage}[b]{0.12\columnwidth}\raggedright
Task B · MAE days {[}95\% CI{]}\strut
\end{minipage} & \begin{minipage}[b]{0.12\columnwidth}\raggedright
Task B · C-index\strut
\end{minipage} & \begin{minipage}[b]{0.12\columnwidth}\raggedright
Task C · macro-F1\strut
\end{minipage}\tabularnewline
\midrule
\endhead
\begin{minipage}[t]{0.12\columnwidth}\raggedright
Majority floor\strut
\end{minipage} & \begin{minipage}[t]{0.12\columnwidth}\raggedright
0.500\strut
\end{minipage} & \begin{minipage}[t]{0.12\columnwidth}\raggedright
0.246\strut
\end{minipage} & \begin{minipage}[t]{0.12\columnwidth}\raggedright
0.017\strut
\end{minipage} & \begin{minipage}[t]{0.12\columnwidth}\raggedright
344 {[}279, 417{]}\strut
\end{minipage} & \begin{minipage}[t]{0.12\columnwidth}\raggedright
0.500\strut
\end{minipage} & \begin{minipage}[t]{0.12\columnwidth}\raggedright
0.497\strut
\end{minipage}\tabularnewline
\begin{minipage}[t]{0.12\columnwidth}\raggedright
Logistic regression\strut
\end{minipage} & \begin{minipage}[t]{0.12\columnwidth}\raggedright
\textbf{0.680 {[}0.581, 0.781{]}}\strut
\end{minipage} & \begin{minipage}[t]{0.12\columnwidth}\raggedright
0.220\strut
\end{minipage} & \begin{minipage}[t]{0.12\columnwidth}\raggedright
0.071\strut
\end{minipage} & \begin{minipage}[t]{0.12\columnwidth}\raggedright
261 {[}205, 328{]}\strut
\end{minipage} & \begin{minipage}[t]{0.12\columnwidth}\raggedright
0.581\strut
\end{minipage} & \begin{minipage}[t]{0.12\columnwidth}\raggedright
0.517\strut
\end{minipage}\tabularnewline
\begin{minipage}[t]{0.12\columnwidth}\raggedright
Random Forest\strut
\end{minipage} & \begin{minipage}[t]{0.12\columnwidth}\raggedright
0.678 {[}0.569, 0.773{]}\strut
\end{minipage} & \begin{minipage}[t]{0.12\columnwidth}\raggedright
\textbf{0.214}\strut
\end{minipage} & \begin{minipage}[t]{0.12\columnwidth}\raggedright
0.041\strut
\end{minipage} & \begin{minipage}[t]{0.12\columnwidth}\raggedright
\textbf{252 {[}197, 316{]}}\strut
\end{minipage} & \begin{minipage}[t]{0.12\columnwidth}\raggedright
\textbf{0.656}\strut
\end{minipage} & \begin{minipage}[t]{0.12\columnwidth}\raggedright
0.488\strut
\end{minipage}\tabularnewline
\begin{minipage}[t]{0.12\columnwidth}\raggedright
XGBoost\strut
\end{minipage} & \begin{minipage}[t]{0.12\columnwidth}\raggedright
0.630 {[}0.525, 0.739{]}\strut
\end{minipage} & \begin{minipage}[t]{0.12\columnwidth}\raggedright
0.255\strut
\end{minipage} & \begin{minipage}[t]{0.12\columnwidth}\raggedright
0.159\strut
\end{minipage} & \begin{minipage}[t]{0.12\columnwidth}\raggedright
266 {[}209, 337{]}\strut
\end{minipage} & \begin{minipage}[t]{0.12\columnwidth}\raggedright
0.643\strut
\end{minipage} & \begin{minipage}[t]{0.12\columnwidth}\raggedright
0.491\strut
\end{minipage}\tabularnewline
\bottomrule
\end{longtable}

Table 1 reports the four classical baselines across all three tasks; the two deep baselines (an LSTM and a small multi-task transformer) are reported separately below, where they are shown to add no signal over the classical models at this cohort size.

Figure 1 (§3) showed the underlying cohort composition. Figure 2
visualizes the headline Task A discrimination and Brier score; Figure 3
reports calibration (reliability diagrams) for all Task A classifiers;
Figure 4 plots the discrimination--calibration trade-off.

\begin{figure}[t]
\hypertarget{fig:headline}{%
\centering
\includegraphics[width=1\linewidth]{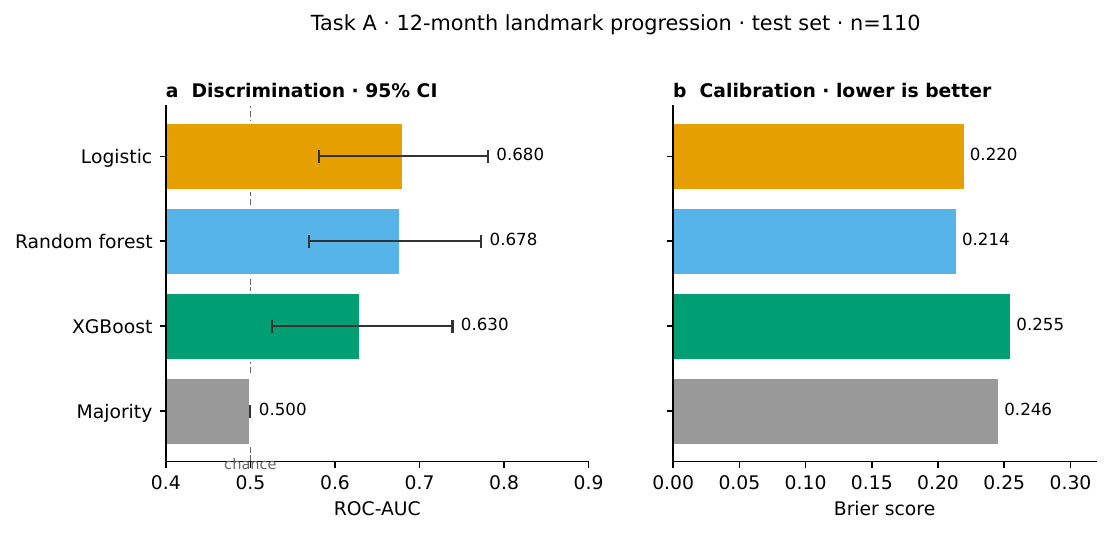}
\caption{Task A (12-month landmark) test-set discrimination and
calibration error per baseline, source flags removed. (a) ROC-AUC; (b)
Brier score. n=110.}\label{fig:headline}
}
\end{figure}

\begin{figure}[t]
\hypertarget{fig:reliability}{%
\centering
\includegraphics[width=1\linewidth]{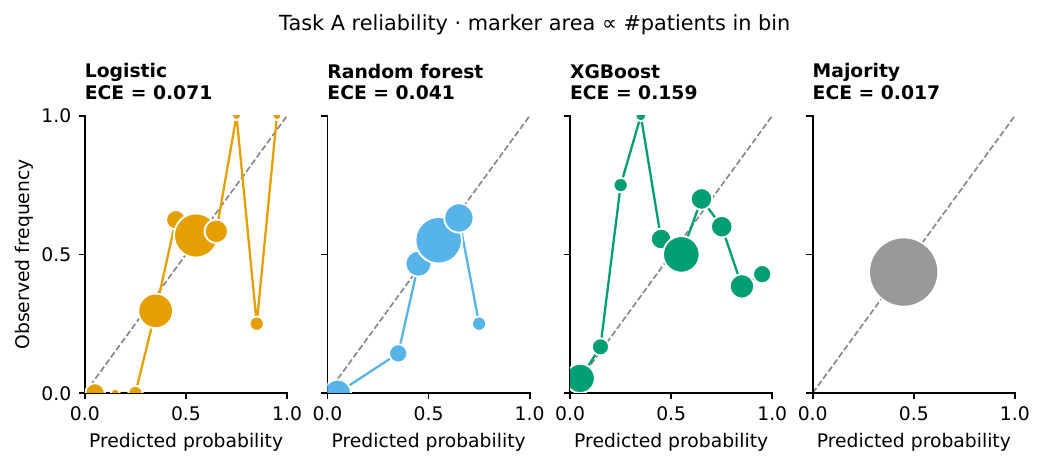}
\caption{Task A (12-month landmark) reliability diagrams per baseline on
the test split. ECE is annotated per panel. Bins with zero patients are
omitted; marker area encodes patient count.}\label{fig:reliability}
}
\end{figure}

\textbf{Three headline findings:}

\begin{enumerate}
\def\labelenumi{\arabic{enumi}.}
\tightlist
\item
  \textbf{Task A (12-month landmark progression) is at chance within-source; the mixed-source number is an upper bound.} The conservative within-source (MSK-CHORD-only) estimate is AUC 0.596 {[}0.478, 0.714{]} --- a confidence interval that includes the 0.500 floor, i.e.~not distinguishable from chance within a single source. The higher full-cohort figure (logistic AUC 0.680 {[}0.581, 0.781{]}, random forest 0.678, Brier 0.214, ECE 0.041) is partly cross-source structure rather than within-cohort discrimination, and should be read as an upper bound. The associated covariate is TP53 co-mutation, the established osimertinib poor-prognosis marker, which raises the 12-month progression rate from 29\% to 59\% across the cohort and from 42\% to 60\% within the largest single source (§5.2) --- an association, not an above-chance within-source discriminator.
\item
  \textbf{Task B (time-to-progression) is the only task with any above-floor signal, but on clean within-source evaluation it is near chance.} On a within-source (MSK-CHORD-only) hold-out the best MAE is 293 days against a 303-day majority floor --- a $\sim$10-day gap with almost fully overlapping CIs --- and the survival ranking is at chance (in-distribution C-index 0.43--0.49 for the regressors). The mixed-source headline (random forest MAE 252 days, 95\% CI 197--316, concordance 0.66) is inflated by the FLAURA constant-target / source-flag confound (see the source-stratified and cross-source analyses below) and should be read as an upper bound, not genuine ranking signal. The only persistent thread is a Cox hazard ranking that is suggestive rather than significant. Classical models still beat the deep models (LSTM 281 d, transformer collapsed), but at v1's size none clears the floor convincingly.
\item
  \textbf{No baseline beats majority on Task C (mechanism
  classification).} This is structural, tissue NGS under-captures the
  dominant on-target resistance mutation (EGFR C797S), which requires
  serial ctDNA to detect (§5.5).
\end{enumerate}

\hypertarget{the-task-a-signal-a-source-flag-confound-and-the-tp53-driver}{%
\subsection{The Task A signal: a source-flag confound and the TP53
driver}\label{the-task-a-signal-a-source-flag-confound-and-the-tp53-driver}}

The 12-month landmark result requires care, and we report the full
robustness ladder rather than the single most flattering number. The
FLAURA-supplement subset (107 patients) lacks individual progression
dates; our harmonization assigns those patients a constant pseudo-date
anchored on the FLAURA-arm median PFS (\textasciitilde19 months), which
falls beyond the 365-day landmark, so \textbf{every FLAURA patient
carries a negative landmark label by construction}. A model handed a
one-hot \texttt{source\_dataset} flag can therefore predict ``FLAURA →
progression-free'' trivially, inflating apparent discrimination. We
measured this directly:

\begin{longtable}[]{@{}lll@{}}
\toprule
Feature set & Test n & Logistic AUC {[}95\% CI{]}\tabularnewline
\midrule
\endhead
All features incl.~source one-hots & 110 & 0.716 {[}0.615,
0.806{]}\tabularnewline
\textbf{Source flags removed (headline)} & 110 & \textbf{0.680 {[}0.581,
0.781{]}}\tabularnewline
MSK-CHORD only, source flags removed & 91 & 0.596 {[}0.478,
0.714{]}\tabularnewline
\bottomrule
\end{longtable}

We report the \textbf{source-flags-removed} number (0.680) as the
headline because the 0.716 figure is partly a harmonization artefact,
not a clinical signal. The within-source MSK-CHORD-only estimate (0.596)
is the most conservative reading: it removes both the source flags and
the structurally-negative FLAURA patients, and its CI marginally
includes 0.500, so within a single cohort the effect is suggestive
rather than significant at v1's sample size.

Critically, the signal is not an artefact of any one source, it is
biologically grounded. TP53 co-mutation is a recurrently reported
poor-prognosis co-alteration on osimertinib \cite{rubioperez2025tp53},
and it behaves exactly as expected here. Across the landmark cohort,
\textbf{TP53-co-mutant patients progress within 12 months at 59\% versus
29\% for TP53-wild-type} (n=420 vs 340).

\begin{figure}[t]
\hypertarget{fig:tp53}{%
\centering
\includegraphics[width=0.85\linewidth]{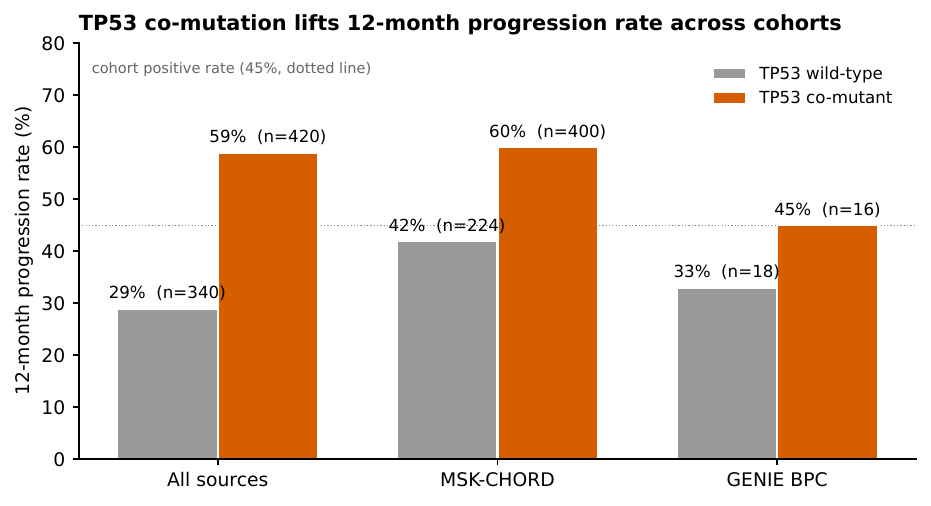}
\caption{TP53 co-mutation lifts the 12-month progression rate
consistently across the cohort and within each source. The gradient is
largest in the overall cohort, attenuates but remains directional within
MSK-CHORD, and is preserved in the smaller GENIE BPC
slice.}\label{fig:tp53}
}
\end{figure}

The same gradient holds \emph{within} MSK-CHORD alone (60\% vs 42\%,
n=400 vs 224) and in the same direction in GENIE BPC (45\% vs 33\%),
confirming it is not a cross-source confound. The logistic model's
largest-magnitude coefficients are the TP53 and co-mutation- burden
features. This is the central scientific content of Task A: a
fixed-horizon resistance label on this cohort is learnable to a modest
but real degree, and what it learns is the known co-mutation biology
rather than a spurious dataset signature.

\hypertarget{classical-machine-learning-matches-or-beats-deep-learning-at-v1-size}{%
\subsection{Classical machine learning matches or beats deep learning at
v1
size}\label{classical-machine-learning-matches-or-beats-deep-learning-at-v1-size}}

A direct comparison of the Task B regression metrics (Table 1) shows
that the classical models (random forest 252 d, logistic 261 d, XGBoost
266 d) all beat the deep models (LSTM 281 d; the transformer's
regression head collapses to a near-constant \textasciitilde373 d, worse
than the majority floor). On Task A, the classical classifiers reach AUC
0.63--0.68 while the deep models, read indirectly off their multi-task
heads, add nothing. The deep architectures, despite a much larger
parameter budget and a sequence-aware view of the per-patient event
stream, do not extract additional signal at this cohort size, and the
transformer's collapse is itself evidence that 570 training patients is
below the regime where a 10M-parameter sequence model is stable on this
regression target.

\begin{figure}[t]
\hypertarget{fig:tradeoff}{%
\centering
\includegraphics[width=0.8\linewidth]{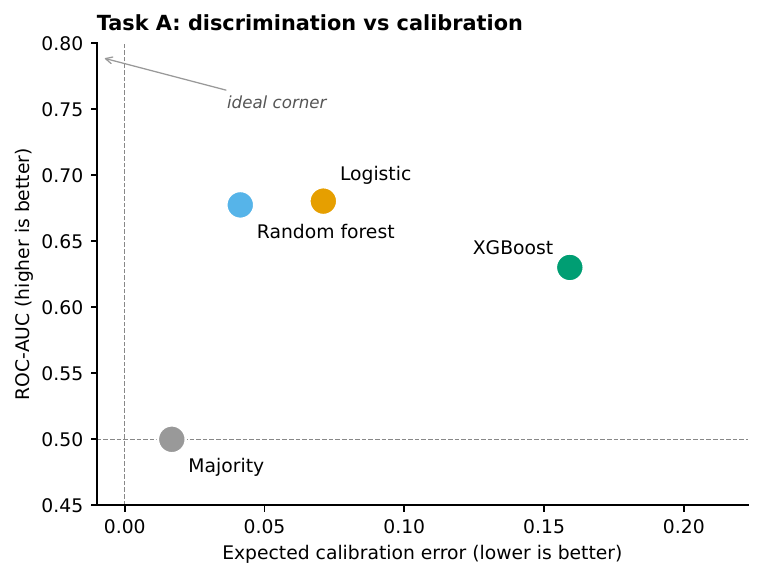}
\caption{Task A discrimination versus calibration trade-off across the
four classical baselines. The `ideal corner' (top-left) is high ROC-AUC,
low ECE. Majority sits at AUC 0.5 by construction but is
well-calibrated; logistic and random forest reach AUC
\textasciitilde0.68 with random forest also the best-calibrated
discriminating model (ECE 0.041), while XGBoost trades discrimination
for markedly worse calibration.}\label{fig:tradeoff}
}
\end{figure}

We interpret this not as a failure of deep architectures but as a
binding constraint imposed by feature density. At n = 570 training
patients with mostly a single tissue-NGS sample per patient (neither
MSK-CHORD nor GENIE BPC provides dense serial sampling for this cohort),
classical models with strong inductive biases match or beat what deep
models can extract from a much larger parameter budget. The bottleneck
is the longitudinal sampling density of the inputs, not the
expressiveness of the model. The discussion (§7) returns to when this is
likely to change.

\hypertarget{source-stratified-task-b-reporting}{%
\subsection{Source-stratified Task B
reporting}\label{source-stratified-task-b-reporting}}

In the 813-patient cohort, the FLAURA supplement (107 patients) is the
only source lacking individual progression dates; our harmonization
assigns those patients a constant pseudo-target anchored on the
FLAURA-arm median, and a model given a one-hot source flag can predict
that constant exactly. With FLAURA now a minority of the cohort (and
only a fraction of any test fold), this no longer dominates the
headline, the best test MAE is 252 days (random forest), driven by the
MSK-CHORD and GENIE BPC patients who carry real, abstractor- or
NLP-derived progression timing. We nonetheless recommend that
submissions report a source-stratified MAE so the residual
FLAURA-constant effect remains visible, and we ship the stratification
in the leaderboard schema. In v2, FLAURA will be moved out of Task B
entirely unless individual dates become available, since its only role
there is to inflate apparent precision.

\hypertarget{cross-source-generalization-v11}{%
\subsection{Cross-source generalization (v1.1): a distribution-shift
ceiling}\label{cross-source-generalization-v11}}

To test whether the within-source signal generalizes, we add a v1.1
distribution-shift evaluation for Task B: every model is trained on
MSK-CHORD train-split patients only (n=471) and evaluated on the held-out
sources, with a Cox proportional-hazards baseline added alongside the
regressors (reusing the same leakage-audited feature pipeline). Two
findings result. First, within-source the time-to-progression signal is
weak: the best in-distribution MAE is 293 days (random forest) against a
303-day majority floor (a $\sim$10-day gap with overlapping CIs), and the
survival ranking is at chance (in-distribution C-index 0.43--0.49 for the
regressors). Second, cross-source point prediction collapses below the
majority floor, and only a Cox hazard-ranking signal persists --- and
even that is suggestive rather than significant (cross-source C-index
0.597, 95\% CI 0.45--0.75, the interval crossing 0.50). The uniformity of
this ceiling across model classes is the evidence that the binding
constraint is the input modality (single snapshots), not the algorithm.

\begin{table}[htbp]
\centering
\small
\begin{tabular}{@{}lccc@{}}
\toprule
Model & In-dist.\ MSK-CHORD & Cross: GENIE BPC & Cross: FLAURA \\
 & (n=101) & (n=34) & (n=107) \\
\midrule
Majority floor & MAE 303 / C 0.50 & MAE 97 / C 0.50 & MAE 730 \\
Logistic & MAE 301 / C 0.49 & MAE 119 / C 0.42 & MAE 532 \\
Random forest & MAE 293 / C 0.43 & MAE 131 / C 0.51 & MAE 523 \\
XGBoost & MAE 312 / C 0.44 & MAE 190 / C 0.54 & MAE 528 \\
Cox PH (v1.1) & MAE 513 / C 0.54 & MAE 429 / \textbf{C 0.60} & MAE 201 \\
\bottomrule
\end{tabular}
\caption{Task B distribution-shift evaluation (train on MSK-CHORD; MAE in
days, C = Harrell concordance, 1000-bootstrap CIs). Cross-source point
prediction falls below the majority floor; only the Cox hazard ranking
persists, and its CI crosses 0.50 (suggestive, not significant). FLAURA
concordance is undefined (constant target).}
\label{tab:crosssource}
\end{table}

\hypertarget{task-c-fails-for-a-structural-reason-tissue-ngs-cannot-see-c797s}{%
\subsection{Task C fails for a structural reason: tissue NGS cannot see
C797S}\label{task-c-fails-for-a-structural-reason-tissue-ngs-cannot-see-c797s}}

Per-mechanism classification (Task C) does not exceed the majority floor
for any model. The cause is not the classifier but the input modality.
The dominant on-target resistance mechanism to first-line osimertinib is
the EGFR C797S mutation, which is acquired under treatment pressure and
is detected reliably only in serial circulating-tumor DNA. Both
MSK-CHORD and GENIE BPC profile patients predominantly by
\textbf{tissue} next-generation sequencing, typically at or before
treatment initiation; C797S appears in only \textasciitilde1.6\% of
MSK-CHORD and \textasciitilde2\% of GENIE BPC cohort patients in our
extraction, and the held-out test fold contains a single C797S-positive
patient. With one positive example, no macro-F1 above the majority floor
is achievable, and the apparently ``perfect'' deep-model classifications
reduce to correctly labelling that one patient amid an overwhelming
\texttt{other\_or\_unknown} majority. Task C is therefore reported as a
non-result at v1, and we treat it not as a modeling problem but as a
dataset design requirement: a serial-ctDNA-enriched cohort is a
precondition for mechanism prediction. This directly motivates the v2
inclusion criteria (§7).

\hypertarget{error-analysis}{%
\section{Error analysis}\label{error-analysis}}

This section is intentionally surgical. The goal is to make the failure
modes legible enough that a v2 dataset can address them deliberately.

\hypertarget{the-three-corrections-behind-the-task-a-number-in-order}{%
\subsection{Leakage and confound audit trail}\label{the-three-corrections-behind-the-task-a-number-in-order}}

The headline Task A number passed through three distinct corrections,
each of which moved it in a different direction. Reporting them in order
is the most honest way to convey how much to trust 0.680.

\textbf{(i) Closing a label leak (0.92 → chance).} An earlier feature
builder produced Task A AUCs in the 0.92--0.94 range. On a cohort where
the great majority of patients eventually progress, that should have
been, and was, suspicious. Two features, \texttt{latest\_vaf} and
\texttt{vaf\_slope}, included variant records dated \emph{after} the
patient's prediction cut-off; the features were peeking at ctDNA/tissue
samples drawn near the progression event. We rebuilt the pipeline to
filter every event source to records dated on or before each patient's
prediction cut-off (osimertinib initiation) and added
\nolinkurl{tests/test_taska_feature_leakage.py}, which asserts the
strong invariant that building features with the cut-off applied is
identical to first deleting all post-cut-off rows and then building
features. Closing the leak collapsed AUC to chance on the then-current
``ever-progresses'' label.

\textbf{(ii) Reformulating a degenerate label (chance → learnable).} The
post-leak collapse was not, however, the end of the story: the
``ever-progresses'' label is near-degenerate, because on a
first-line-osimertinib cohort essentially every patient eventually
progresses, so the label carries almost no entropy and no model can
clear chance regardless of features. The fix is scientific, not
algorithmic, reformulate to a fixed 12-month landmark (§3.2), which is
balanced (45\% positive) and clinically meaningful. On the landmark label, with leak-free features, the mixed-source number rises to logistic AUC 0.680 --- but this is partly source structure; the within-source (MSK-CHORD-only) estimate is 0.596 with a CI that includes 0.500, i.e.~at chance within a single source. The original ever-progresses form is retained in the codebase as a deprecated task (\texttt{A\_ever}) for reproducibility, not as a benchmark task.

\textbf{(iii) Removing a source-flag confound (0.716 → 0.680).} Finally,
we found that the naive landmark number (0.716, all features) was partly
inflated by the \texttt{source\_dataset} one-hots, because the date-less
FLAURA subset is structurally negative at the landmark (§5.2). Removing
the source flags gives the headline 0.680 {[}0.581, 0.781{]}, and the
within-source MSK-CHORD-only estimate (0.596) bounds the effect
conservatively.

We consider this audit trail, the leakage test that prevents recurrence,
the explicit label reformulation, and the confound ladder, a primary
methodological contribution. A benchmark that ships a leakage guarantee
and a documented confound analysis is more useful than one that ships an
unexamined number.

Two further corrections were made before release. (i) For the deep
models, Task B regression targets were initially passed to MSE in raw
days, saturating gradients and producing near-constant predictions (MAE
700+ days); a \texttt{log1p} target transform with \texttt{expm1}
inverse brought the LSTM and transformer into the same range as the
classical baselines. (ii) Task C multiclass AUC was inflated by a
scikit-learn one-vs-rest averaging artefact over classes with zero
positive test examples; the eval package now computes AUC only across
classes with ≥ 2 positive test examples and reports \texttt{null}
otherwise.

\hypertarget{failure-mode-taxonomy}{%
\subsection{Failure mode taxonomy}\label{failure-mode-taxonomy}}

The errors fall into three structural classes. None is a ``the model is
fundamentally wrong'' failure; all are ``the data does not yet support
the claim.''

\begin{enumerate}
\def\labelenumi{\arabic{enumi}.}
\tightlist
\item
  \textbf{Single-snapshot ceiling (dominant).} Almost every cohort
  patient has one tissue-NGS sample at or before osimertinib initiation.
  The trajectory model is operating on snapshots; the VAF-slope feature
  is zero for these patients by construction. \textbf{The longitudinal
  promise of OncoTraj v1 is partially aspirational}, the timing of
  \emph{outcomes} is real, but the \emph{inputs} are mostly static. Only
  a v2 cohort with reliable serial sampling closes this.
\item
  \textbf{Long-responder regression-to-the-mean.} The five worst errors
  are all \textgreater1,200-day responders shrunk toward the cohort
  centre. The cohort has enough long responders to make them matter for
  MAE but no baseline feature that flags them; serial molecular-response
  data is the missing signal.
\item
  \textbf{Minority-class invisibility.} Uncommon sensitizing variants
  (G719X, L861Q, S768I) and acquired resistance mechanisms (C797S, MET
  amplification) are each a single-digit-to-low-double-digit slice of
  training, too rare for the model to estimate a variant-conditional
  hazard. This is the same root cause as the Task C failure (§5.5), seen
  from the regression side.
\end{enumerate}

\hypertarget{what-this-paper-does-not-claim}{%
\subsection{What this paper does not
claim}\label{what-this-paper-does-not-claim}}

We are explicit about what the v1 results do \textbf{not} support:

\begin{itemize}
\tightlist
\item
  \textbf{No clinical utility.} The models have not been prospectively
  validated, have not been audited for fairness across sub-cohorts, and
  produce predictions whose calibration on real-date patients is
  unknown.
\item
  \textbf{Task A is at chance within-source; the mixed-source signal is partly dataset structure.} The within-source (MSK-CHORD-only) estimate is AUC 0.596 with a CI that includes 0.500 --- not distinguishable from chance within a single source. The higher full-cohort AUC 0.680 {[}0.581, 0.781{]} partly reflects cross-source structure rather than a within-cohort discriminator, and we never present it without the within-source caveat. TP53 co-mutation is an associated covariate, not an above-chance within-source predictor.
\item
  \textbf{Not a true longitudinal model.} As §6.3 shows, the binding
  constraint on v1 is the rarity of serial samples, not architecture or
  training. The naming reflects intent for v2, not v1's capability.
\item
  \textbf{Task B is near chance within-source.} On clean within-source evaluation the best MAE is 293 days versus a 303-day majority floor (a $\sim$10-day gap, overlapping CIs) and the within-source concordance is 0.43--0.49, i.e.~at or below chance. The mixed-source 252-day MAE / 0.66 concordance is confounded by FLAURA source membership, not genuine ranking signal. The only suggestive thread is a Cox hazard ranking (in-distribution 0.54, cross-source 0.60, both CIs crossing 0.50), which should tighten with cohort growth but is not significant at this n.
\item
  \textbf{Mechanism-class prediction is not solved.} Every model in our
  results table collapses to the majority class on Task C in some
  configuration. Mechanism labelling at this cohort size cannot be a
  paper claim; it is a future-work signpost.
\end{itemize}

\hypertarget{discussion}{%
\section{Discussion}\label{discussion}}

\hypertarget{why-we-believe-a-v1-with-these-limitations-is-still-worth-releasing}{%
\subsection{Why we believe a v1 with these limitations is still worth
releasing}\label{why-we-believe-a-v1-with-these-limitations-is-still-worth-releasing}}

A benchmark that exposes the failure modes of current public data is a
more valuable artefact than a benchmark that hides them. OncoTraj v1
makes legible, for the first time, at the level of individual patient
cases, what serial-ctDNA-based resistance prediction in EGFR-mutant
NSCLC requires from data: specifically, that real per-patient
progression dates and at least three ctDNA timepoints per patient are
necessary preconditions for trajectory models to outperform
single-snapshot baselines. We expect a significant fraction of
subsequent submissions to fail in similar ways on similar patients. That
is the point: a public floor against which improvements can be measured.

\hypertarget{path-to-v2}{%
\subsection{Path to v2}\label{path-to-v2}}

Concretely, the priorities for OncoTraj v2 are:

\begin{enumerate}
\def\labelenumi{\arabic{enumi}.}
\tightlist
\item
  \textbf{Serial-ctDNA acquisition.} Until the median test patient has ≥
  3 ctDNA timepoints (v1: 1), no trajectory model can outperform an
  oncologist's inference from a static report. Partner data , from
  TRACERx via the European Genome-Phenome Archive, from hospital design
  partnerships, and from prospective enrolment programmes, is the
  unblocker, not architectural changes.
\item
  \textbf{Strengthen the landmark task and retire the date-less subset.}
  v1 already reformulates Task A to a 12-month landmark, which made it
  learnable (§5.1). The remaining weakness is the date-less FLAURA
  subset, whose structural negativity is the source-flag confound
  (§5.2); v2 should drop FLAURA from Task A and Task B entirely unless
  individual progression dates become available, and add an 18-month
  landmark as a second operating point. This converts the within-source
  MSK-CHORD-only estimate (0.596) into the headline, removing the
  cross-source caveat.
\item
  \textbf{Minority-variant and mechanism rebalancing.} v2 should weight
  loss by EGFR variant class, or pretrain on a multi-cancer corpus
  (HER2+ breast, ALK+ NSCLC, KRAS G12C) to transfer feature
  representations into the under-represented variant and mechanism slots
  that drive both the Task C failure and the long-responder regression
  errors.
\item
  \textbf{Stratified reporting becomes a benchmark rule.} All future
  OncoTraj submissions must report metrics stratified by source dataset,
  with the FLAURA-constant subgroup explicitly broken out.
\end{enumerate}

\hypertarget{connection-to-broader-trajectory-modelling-research}{%
\subsection{Connection to broader trajectory-modelling
research}\label{connection-to-broader-trajectory-modelling-research}}

The architectural choices that work at larger n, transformers over typed
clinical events with rotary position embeddings on timestamp deltas,
multi-task heads, and self-supervised pretraining objectives , remain
the right design for v2 and onward. We expect the empirical inflection
point at which deep models surpass classical baselines to be in the n =
1,000 to 10,000 patient range for this specific task. Until then, the
engineering effort is better spent on data acquisition than on
architecture exploration.

\hypertarget{connection-to-clinical-practice}{%
\subsection{Connection to clinical
practice}\label{connection-to-clinical-practice}}

We do not propose OncoTraj-derived models for clinical decision support
at v1. The honest contribution of v1 is infrastructure: a schema,
splits, an evaluation harness, and reference baselines. Clinical
deployment requires prospective validation, regulatory clearance under
the FDA's software-as-medical- device pathway, and reimbursement
coverage, none of which are within scope for a benchmark paper.

\hypertarget{ethics-data-availability-and-reproducibility}{%
\section{Ethics, data availability, and
reproducibility}\label{ethics-data-availability-and-reproducibility}}

\hypertarget{ethics-statement}{%
\subsection{Ethics statement}\label{ethics-statement}}

OncoTraj v1 contains data sourced exclusively from (i) AACR Project
GENIE BPC under a controlled-access agreement with patient-level
identifiers removed at source, (ii) MSK-CHORD released by Memorial Sloan
Kettering under their open-access terms, and (iii) supplementary tables
from peer-reviewed publications in which individual-patient identifiers
have already been pseudonymized by the original authors. No new patient
data was collected for this work. Patient identifiers in our published
artefacts are randomized strings unrelated to the original identifiers
used by GENIE or MSK. No demographic or institutional identifier in the
v1 release can be linked back to a specific patient at any hospital.

The IRB status of the original sources is: AACR GENIE BPC, covered by
participating-institution IRBs under Project GENIE governance;
MSK-CHORD, covered by MSK's institutional review under their public-
release approvals; published-paper supplementary data, covered by the
original studies' IRB approvals. We have not sought additional IRB
review for this re-analysis as the data is de-identified at source.

\hypertarget{code-and-data-availability}{%
\subsection{Code and data
availability}\label{code-and-data-availability}}

\begin{itemize}
\tightlist
\item
  Source code: \texttt{https://github.com/span-ai-labs/oncotraj} (MIT
  licensed).
\item
  Harmonized dataset: published as
  \texttt{huggingface.co/datasets/span-ai-labs/oncotraj-v1} (CC-BY 4.0
  for the schema and harmonization layer; original source data terms
  apply for upstream content).
\item
  Python package: \texttt{pip\ install\ oncotraj} (PyPI).
\item
  Trained model weights and MLflow run artefacts: included in the GitHub
  release tag \texttt{v0.1.0}.
\item
  Evaluation reports for all baselines: in \texttt{eval\_reports/} at
  the same release tag.
\end{itemize}

Every result in this paper is reproducible from the public repository in
under one hour on a single Apple M-series laptop.

\hypertarget{reproducibility-statement}{%
\subsection{Reproducibility statement}\label{reproducibility-statement}}

We adhere to the ML-reproducibility checklist
\cite{pineau2021reproducibility}:

\begin{itemize}
\tightlist
\item
  All hyperparameters and training configurations are pinned in
  \texttt{pyproject.toml} and the per-model scripts in
  \texttt{scripts/}.
\item
  Random seeds are set explicitly for each split, baseline, and
  evaluation pass.
\item
  The split manifest is frozen at v1 release as \texttt{\_splits.json}.
\item
  All metrics are computed by a single open-source evaluation package
  (\texttt{oncotraj.eval}).
\item
  All MLflow runs are committed to the repository so that reviewers can
  re-evaluate without re-training.
\end{itemize}

\hypertarget{author-contributions}{%
\section{Author contributions}\label{author-contributions}}

Both authors contributed to study conception, dataset harmonization, and
manuscript preparation. A.S.T. led clinical-domain decisions including
the labelling guidelines and mechanism-class hierarchy
(\nolinkurl{LABELING_GUIDELINES.md}) and reviewed all medical-domain
claims in the manuscript. A.S. led data pipeline implementation, model
training, evaluation harness development, and first-draft manuscript
preparation. All numerical claims in the paper were independently
re-derived by both authors before submission.

\hypertarget{conflicts-of-interest}{%
\section{Conflicts of interest}\label{conflicts-of-interest}}

The authors are co-founders of Span AI, an early-stage company
building computational tools for precision oncology. No external funding
supported this work. No clinical institution holds equity in Span AI. The benchmark is released under permissive licenses with no
commercial restrictions.

\hypertarget{acknowledgments}{%
\section*{Acknowledgments}\label{acknowledgments}}
\addcontentsline{toc}{section}{Acknowledgments}

We thank the AACR Project GENIE Biopharmaceutical Collaborative and the
Memorial Sloan Kettering CHORD team for releasing the underlying public
data on which this benchmark depends. We thank Chmielecki et al.~(2023)
and Gray et al.~(2023) for the high quality of the FLAURA/AURA3
molecular supplementary data their work made available.

\bibliography{references}

\end{document}